\newif\ifarxiv
\arxivtrue 

\ifarxiv
\documentclass[12pt]{article}
	\usepackage[margin=1in]{geometry}	
	\usepackage[slantedGreek]{mathpazo}
	\usepackage[pdftex]{graphicx}
	\usepackage{amsfonts}
	\usepackage{amssymb}
	\usepackage{amsthm}
	\usepackage{graphicx,float}
	\usepackage{amsmath}%
	\usepackage{hyperref}
	\usepackage{natbib}
	\usepackage{color,soul,marginnote}
\else
\fi

\usepackage{tabularray}
\usepackage{float}
\usepackage{graphicx}
\UseTblrLibrary{booktabs}
\UseTblrLibrary{siunitx}

\NewTableCommand{\tinytableDefineColor}[3]{\definecolor{#1}{#2}{#3}}

\newcommand{\defeq}{:=}

\newcommand{\vnorm}[1]{\left|\left|#1\right|\right|}

\ifarxiv
	\title{Kolmogorov GAM Networks  are all you need!}
	\author{	
	\makebox[.4\linewidth]{Sarah Polson}\\\textit{  Math Institute}\\\textit{  Oxford University}\\\and 
	\makebox[.4\linewidth]{Vadim Sokolov\footnote{Sarah Polson is at the Maths  Institute, University of Oxford. Vadim Sokolov is Associate Professor at Volgenau School of Engineering, George Mason University, USA: vsokolov@gmu.org.}}\\\textit{ Department of Systems Engineering }\\\textit{  and Operations Research}\\\textit{ George Mason University}
	}
	\date{First Draft: November 1, 2024\\This Draft: \today}
\else
\fi

\graphicspath{{./fig/}{./code/fig/}}

\begin{document}
\ifarxiv
\maketitle
\begin{abstract}
\noindent	Kolmogorov GAM (K-GAM) networks are shown to be an efficient architecture for training and inference.  They are an additive model with an embedding that is independent of the function of interest. They provide an alternative to the transformer architecture. They are the machine learning version of  Kolmogorov's Superposition Theorem (KST)  which provides an efficient representations of a multivariate function. Such representations have use in machine learning for encoding dictionaries (a.k.a. "look-up" tables).  KST theory also provides a representation based on translates of the K\"oppen function. The goal of our paper is to interpret this representation in a machine learning context for applications in Artificial Intelligence (AI). Our architecture is equivalent to a topological embedding which is independent  of the function together with an additive layer that uses a Generalized Additive Model (GAM). This provides a class of learning procedures with far fewer parameters than current deep learning algorithms. Implementation  can be parallelizable which makes our algorithms computationally attractive. To illustrate our methodology, we use the Iris data from statistical learning. We also show that our additive model with non-linear embedding provides an alternative to  transformer architectures which from a statistical viewpoint are  kernel smoothers. Additive KAN models therefore provide a natural alternative to transformers. Finally, we conclude with directions for future research.
\end{abstract}
\else
\fi


\newpage

\section{Introduction}

The landscape of modern machine learning has been shaped by the exponential growth in computational power, particularly through advances in GPU technology and frameworks like PyTorch. While Moore's Law has continued to drive hardware improvements and CUDA algorithms have revolutionized our ability to process vast amounts of internet data, we pose the following question: can we achieve superior performance through mathematically efficient representations of multivariate functions rather than raw computational power?

In this paper, we introduce Kolmogorov generalized additive models (K-GAM), a novel neural network architecture whose additive structure enables both simplified training and accelerated inference through a marked reduction in parameters compared to traditional approaches. K-GAMs leverage Kolmogorov's Superposition Theorem by utilizing a composition of two key components: a universal K\"oppen embedding (functioning as a space-filling curve), followed by a trainable outer function $g: [0,1] \rightarrow [0,1]$. Unlike the original iterative look-up table approach proposed by K\"oppen and Sprecher, which faces NP-hard computational challenges, we implement the outer function using a ReLU neural network that can be efficiently trained using standard optimization techniques.

A fundamental challenge in machine learning lies in effectively handling high-dimensional input-output relationships. This challenge manifests itself in two distinct but related tasks. First, one task is to construct a ``look-up'' table (dictionary) for fast search and retrieval of input-output examples. This is an encoding and can be thought of as a data compression problem. Second, and perhaps more importantly, we must develop prediction rules that can generalize beyond these examples to handle arbitrary inputs.

More formally, we seek to find a good predictor function $f(x)$ that maps an input $x$ to its output prediction $y$. In practice, the input $x$ is typically a high-dimensional vector:
\[
y = f ( x )  \; \; {\rm where}  \; \; x =  ( x_1 , \ldots , x_d ) 
\]
Given  a training dataset $(y_i,x_i)_{i=1}^N$ of example input-output pairs, our goal is to train a model, i.e. to find the function $f$. The key question is: \emph{how do we represent a multivariate function so as to obtain a desirable $f$?}

We demonstrate that transformer architectures can be decomposed into two fundamental operations: a particular form of embedding followed by kernel smoothing. Hence, K-GAM models are natural competitors to transformers. The significance of this connection becomes clear when we consider that transformers have largely supplanted alternative architectures across a wide range of applications. Previous work in this direction - Bryant and Leni - demonstrated the potential of K\"oppen-based approaches only for $ 2 \times 2 $ image reconstruction tasks. Our analysis extends this framework to encompass a wider spectrum of machine learning problems.

The rest of the paper is outlined as follows. Section 2 discusses the Kolmogorov Superposition Theorem. Section 3 introduces Kolmogorov generalized additive models (K-GAM). Section 4 describes transformer architectures as kernel smoothers. Section 5 provides an application of GAM-Kolmogorov embeddings to the Iris dataset. Section 6 concludes with directions for future research.

\section{Kolmogorov Superposition Theorem (KST)}
Kolmogorov demonstrated that any real-valued continuous function $f({\bf x})$ defined on $E^n$ can be represented as a convolution of two single variable functions:
\[
f(x_1,\ldots,x_n) = \sum_{q=1}^{2n+1} g_q\left(\phi_q(x_1,\ldots,x_n)\right)
\]
where $g_q$ are continuous single-variable functions defined on $\phi_q(E^n)$. Kolmogorov further showed that the $\phi_q$ functions can be decomposed into sums of single-variable functions:
\[
\phi_q(x_1,\ldots,x_n) = \sum_{i=1}^n \psi_{q,i}(x_i)
\]
This result is known as Kolmogorov representation theorem \cite{kolmogorov1956representation} and is often written in the following form:
\[
f(x_1,\ldots,x_n) = \sum_{q=1}^{2n+1} g_q\left(\sum_{i=1}^n \psi_{q,i}(x_i)\right)
\]
\\
Our goal is to show that it applies directly to the problems of machine learning. Furthermore, we interpret the architecture as a statistical model which allows us to draw on uncertainty quantification and approximation bounds. To understand the power of this representation from a machine learning perspective, we can think of $\psi_{q,i}(x_i)$ as features. Note that the number of features is then $2n+1$ for an $n$-dimensional input. This representation is also remarkable in the sense that $ \Psi_q (x) $ can be viewed as an embedding which is \emph{independent} of $f$.

\subsection{Inner and Outer Functions}

The theorem has seen several refinements over time. \cite{sprecher1965structure} made a crucial advancement by proving that the inner functions could be H\"older continuous. \cite{fridman1967improvement} later strengthened this result, showing that these functions could be chosen to be Lipschitz continuous, though this required modifications to both the outer and inner functions.

We focus on the Kolmogorov-Sprecher formulation, as modified by \cite{koppen2002training,koppen2000curse} and \cite{braun2009application}, which expresses any multivariate continuous function as:
$$
f( x_1 , \ldots , x_d ) = \sum_{q=0}^{2d} G \circ \Psi_q ( x ) \; \; {\rm where} \; \; \Psi_q ( x) = \sum_{p=1}^d \alpha_p \psi ( x_p + q \alpha ) + \delta_q
$$
Here, the inner functions $\Psi_q$ are constructed as sums of translated versions of a single univariate function $\psi$, known as the K\"oppen function. The K\"oppen function is monotone, H\"older continuous with smoothness parameters, and has fractal-like properties. Recent work by \cite{montanelli2020error} provides a modern construction of the K\"oppen function and establishes theoretical approximation results with ReLU networks. \cite{schmidt-hieber2020nonparametric} extends these these theoretical results and shows that a function with smaller number of parameters can be used. He uses B-adic representation, where
\[
f(x_1,\ldots,x_d) = G\left(\sum_{p=1}^{d}B^{-p}\psi(x_p)\right),
\]
The structure of $\Psi_q$ differs fundamentally from the approaches used in traditional neural networks, which rely on hyperplanes (as in ReLU networks) or decision trees. In $\Psi_q$, a slice in any given univariate direction cuts the regions defined by the embedding in half.

The inner functions $\Psi_q$ partition the input space into distinct regions, and the outer function, $g$, must be constructed to provide the correct output values across the regions that the inner function defines. The outer function, $g$, can be determined via a computationally intensive process of averaging. For each input configuration, the inner functions $\Psi_q$ generate a unique encoding, and $g$ must map this encoding to the appropriate value of $f(x)$. This creates a dictionary-like structure that associates each region with its corresponding output value. K\"oppen made significant contributions by correcting Sprecher's original proof of this construction process, with improvements to the computational algorithm later suggested by \cite{actor2018computation} and \cite{demb2021note}. Braun further enhanced the understanding by providing precise definitions of the shift parameters $\delta_k$ and characterizing the topological structure induced by $\Psi_q$.

A fundamental trade-off in KST exists between function smoothness and dimensionality. The inner functions $\psi_{p,q}$ can be chosen from two different function spaces, each offering distinct advantages. The first option is to use functions from $C^1([0,1])$, but this limits the network's ability to handle higher dimensions effectively. The second option is to relax the smoothness requirement to H\"older continuous functions ($\psi_{p,q} \in \text{Holder}_\alpha([0,1])$), which satisfy the inequality $|\psi(x) - \psi(y)| < |x-y|^\alpha$. These functions are less smooth, but this ``roughness'' enables better approximation in higher dimensions.

\subsection{Ridge and Projection Pursuit Regression}
To understand the significance of this trade-off, we consider ridge functions, which represent a fundamental building block in multivariate analysis. Since our ultimate goal is to model arbitrary multivariate functions $f$, we need a way to reduce dimensionality while preserving the ability to capture nonlinear relationships. Ridge functions accomplish this by representing one of the simplest forms of nonlinear multivariate functions, requiring only a single linear projection and a univariate nonlinear transformation. Formally, a ridge function $f: \mathbb{R}^n \rightarrow \mathbb{R}$ takes the form $f(x) = g(w^Tx)$, where $g$ is a univariate function and $x,w \in \mathbb{R}^n$. The non-zero vector $w$ is called the direction. The term "ridge" reflects a key geometric property: the function remains constant along any direction orthogonal to $w$. Specifically, for any direction $u$ such that $w^Tu = 0$, we have
\[
f(x+u) = g(w^T(x+u)) = g(w^Tx) = f(x)
\]
This structural simplicity makes ridge functions particularly useful as building blocks for high-dimensional approximation. \cite{klusowski2016risk} provided bounds for the approximation error when a function with finite spectral norm is approximated by ridge functions.

Ridge functions play a central role in high-dimensional statistical analysis. For example, projection pursuit regression approximates input-output relations using a linear combination of ridge functions \citep{friedman1981projection,huber1985projection,jones1992}:
\[
\phi(x) = \sum_{i=1}^{p}g_i(w_i^Tx),
\]
where both the directions $w_i$ and functions $g_i$ are variables and $w_i^Tx$ are one-dimensional projections of the input vector. The vector $w_i^Tx$ is a projection of the input vector $x$ onto a one-dimensional space and $g_i(w_i^Tx)$ can be though as a feature calculated from data. \cite{diaconis1984nonlinear} use nonlinear functions of linear combinations, laying important groundwork for deep learning.

\subsection{Kolmogorov-Arnold Networks}

A significant development has been the emergence of Kolmogorov-Arnold Networks (KANs). The key innovation of KANs is their use of learnable functions rather than weights on the network edges. This replaces traditional linear weights with univariate functions, typically parametrized by splines, enhancing both representational capacity and interpretability.

\cite{hecht-nielsen1987kolmogorovs} established the first practical connection between KST and neural networks by showing that any KAN can be constructed as a 3-layer MLP. \cite{liu2024kana} and \cite{polar2020deep} consider KST in the form of sums of functions, a two layer model:
$$
f( x_1 , \ldots , x_d ) = f( x) = ( g \circ \psi ) (x ) 
$$
Polar (2023) shows KAN not only a superposition of functions but also a particular case of a tree of discrete Urysohn operators:
$$
U(x_1 , \ldots , x_d ) = \sum_{j=1}^d g_j (x_j ) 
$$
This insight leads to a fast scalable algorithm that avoids back-propagation, applicable to any GAM model, using a projection descent method with a Newton-Kacmarz scheme.

Recent research has explored various functional classes for the outer functions, including Wan-KAN \cite{bozorgasl2024wavkan}, SinceKAN \cite{reinhardt2024sinekan}, and functional combinations \cite{ta2024fckan}. \cite{liu2024kan} demonstrated that KAN networks can outperform traditional MLPs, while \cite{lai2024optimal} proved their optimality in terms of approximation error\footnote{Any real-valued continuous function defined on $ [0,1]^d$ can be represented as
$f(x_1,\ldots,x_d) = \sum_{q=0}^{2d} g_q \left( \phi_q(x_1,\ldots,x_d)\right) $.
With continuous single-variable function $g_q$ defined on $\phi_q(E^n)$.
Given $2n+1$ continuous functions of $n$ variables defined on $[0,1]^d$. 
$y_q = \phi_q(x_1,\ldots,x_d), \quad q = 0,1,\ldots,2d $
with property
$\phi_q(S_{q,k}) \cap \phi_{q'}(S_{q,k'}) = \emptyset, \quad k\ne k', \quad q, = 0,1,\ldots,2d $.
For any integer $n\ge 2$, let $S_{q,k}$ be a family of mutually disjoint $n$-dimensional cubes
$S_{q,k} \cap S_{q,k'} = \emptyset, \quad k\ne k', \quad k,k'  = 1,2,\ldots, q = 1,2,\ldots,2n+1 $.
The diameter of each cube is approaches zero as $k\to\infty$ and every point of a unit cube $ [0,1]^n$ belongs to at least $n+1$ cubes $S_{q,k}$ for each $k$. }.

Early theoretical groundwork for adaptive learning in KANs was laid by \cite{hecht-nielsen1987counterpropagation}, who introduced the concept of counter-propagation networks where outer functions self-organize in response to input-output pairs $(x,y)$. Theoretical understanding has continued to advance, with \cite{ismayilova2023kolmogorov} showing that Kolmogorov networks with two hidden layers can precisely represent continuous, discontinuous, and unbounded multivariate functions, depending on the activation function choice in the second layer. \cite{ismailov2022three} extended this to three-layer networks for discontinuous functions.

\section{Kolmogorov Generalized Additive Models (K-GAM)}

We take a different approach. Rather than using learnable functions as network nodes activations, we directly use KST representation. This is a 2-layer network with a non-differentiable inner function. The network's architecture can be expressed as:
$$
f(x_1,\ldots,x_d) = \sum_{q=0}^{2d} g_q(z_q)
$$
where the inner layer performs an embedding from $[0,1]^d$ to $\mathbb{R}^{2d+1}$ via:
$$
z_q = \eta_q ( x_1 , \ldots , x_d ) = \sum_{p=1}^ d \lambda_p \psi  ( x_p + q a ) 
$$
Here, $\lambda_p = \sum_{r=1}^\infty \gamma^{-(p-1)\beta(r)}$ is a $p$-adic expansion with $\beta(r) = (n^r-1)/(n-1) $ and  $\gamma \geq d+2$ with $a = (\gamma(\gamma-1))^{-1}$.

The K\"oppen function $\psi$ is defined through a recursive limit:
\[
\psi(x) = \lim_{k \rightarrow \infty} \psi_k\left(\sum_{l=1}^{k}i_l\gamma^{-l}\right)
\]
where each $x \in [0,1]$ has the representation:
\[
x = \sum_{l=1}^{\infty}i_l\gamma^{-l} = \lim_{k \rightarrow \infty} \left(\sum_{l=1}^{k}i_l\gamma^{-l}\right)
\]
and $\psi_k$ is defined recursively as:
\[
\psi_k = 
\begin{cases}
    d, & d \in D_1\\
    \psi_{k-1}(d-i_k\gamma^{-k}) + i_k\gamma^{-\beta_n(k)}, & d \in D_k,k>1,i_k<\gamma-1\\
    \frac{1}{2}\left(\psi_k(d-\gamma^{-k}) + \psi_{k-1}(d+\gamma^{-k})\right), & d \in D_k, k>1, i_k = \gamma - 1
\end{cases}
\]

\begin{figure}
\begin{tabular}{cc}
	\includegraphics[width=0.45\textwidth]{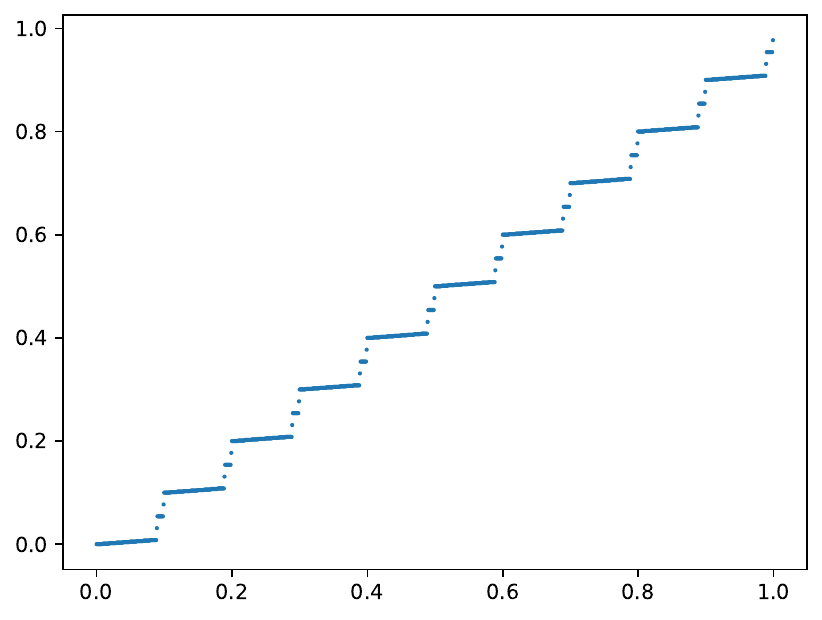} & \includegraphics[width=0.45\textwidth]{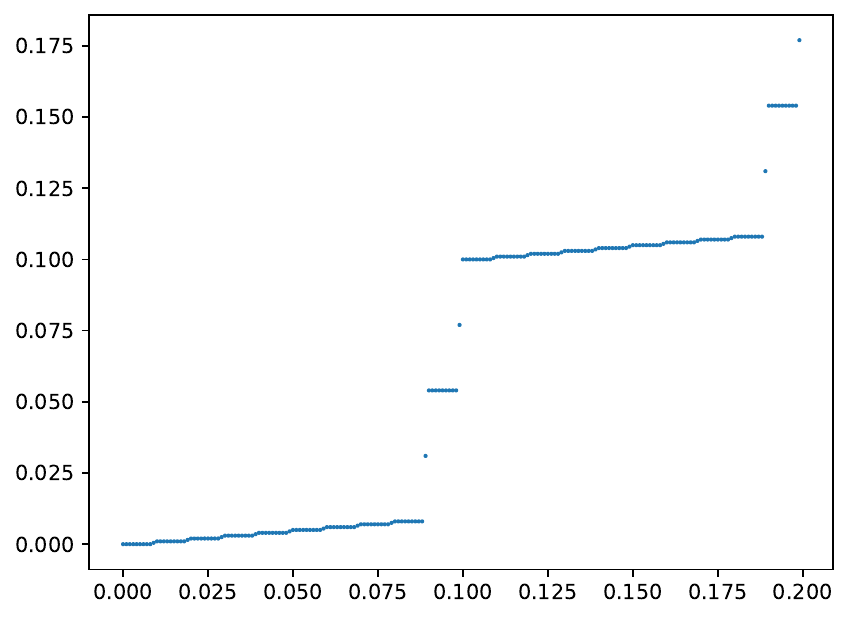}\\
	(a) [0,1], $k=3$ & (b) [0,0.2], $k=3$\\
	\includegraphics[width=0.45\textwidth]{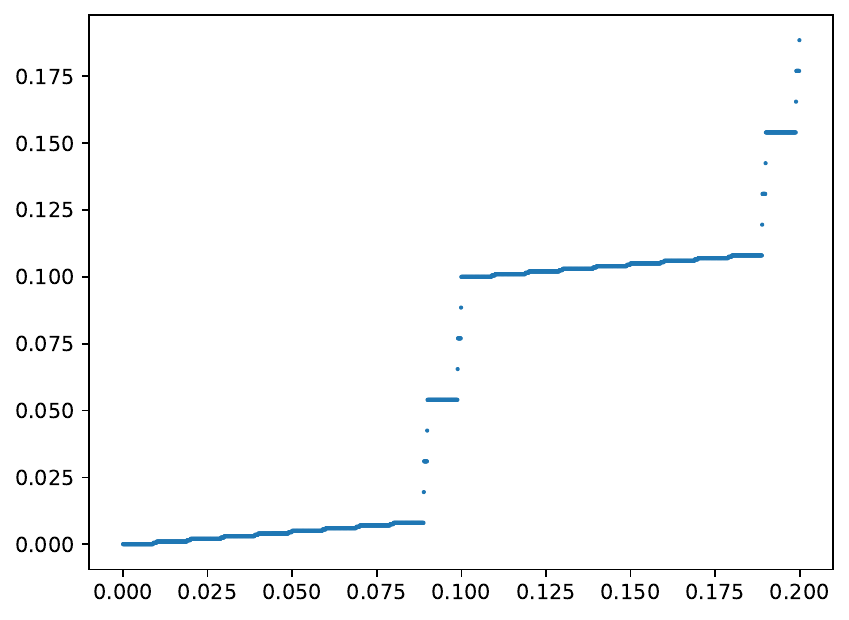} & \includegraphics[width=0.45\textwidth]{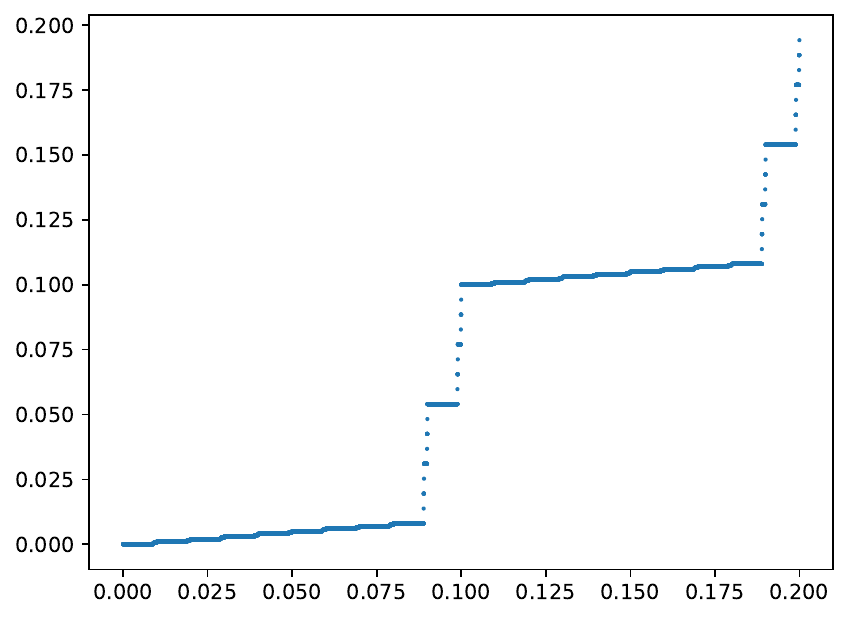}\\
	(a) [0,0.2], $k=4$ & (b) [0,0.2], $k=5$\\
\end{tabular}
\caption{K\"oppen function $\psi_k$ for $k=3,4,5$, $\gamma=10$}\label{fig:Koppen}
\end{figure}

Figure \ref{fig:Koppen} shows the plot of the K\"oppen function $\psi_k$ for $k=3$ on interval $[0,1]$ and $[0,0.2]$ (top row). The bottom row shows a ``zoomed-in'' view of the function for $k=4$ and $k=5$ on the interval $[0,0.2]$. The function has a fractal-like structure, with the number of discontinuities increasing with $k$. The K\"oppen function is a key component of the KST representation, providing a topological embedding of the input space that is independent of the target function $f$. This embedding is crucial for the network's ability to capture complex multivariate relationships.

The most striking aspect of KST is that it leads to a Generalized Additive Model (GAM) with fixed features that are independent of the target function $f$. These features, determined by the K\"oppen function, provide universal topological information about the input space, effectively implementing a k-nearest neighbors structure that is inherent to the representation. The outer function $g$ is then responsible for learning the relationship between these features and the target function $f$. This separation of feature engineering and learning is a key advantage of K-GAM networks, enabling efficient training and inference.

\paragraph{Theorem (K-GAM)} Any function and dataset can be represented as a GAM with feature engineering (topological information) given by features $z_k$ in the hidden layer:
\begin{align*}
y_i &= \sum_{k=1}^{2n+1} g(z_k)\\
z_k &= \sum_{j=1}^n \lambda^k\psi(x_j + \epsilon k) + k
\end{align*}
where $\psi$ is a single activation function common to all nodes, and $g$ is a single outer function.

\paragraph{Proof} Take the Sprecher-K\"oppen representation for $ y=f(x)$ and repeat $n$ times for $ y_i = f ( x_i ) $. As in \cite{sprecher1972improvement}, we can decompose with Lipschitz continuous functions:
$$
f(x_1,\ldots,x_n) = \sum_{j=0}^{2p} \phi_j\left(\sum_{i=1}^n \lambda_i\phi(x_i + ja)\right)
$$
where $\lambda_1 = 1 > \lambda_2 \ldots > \lambda_n$ and $a = \frac{1}{(2n+1)(2n+2)}$. Following \cite{braun2009application}, we can choose $\lambda_k = \lambda^k$ for some $\lambda < 1$.

Now, for a dataset $\{(x_i, y_i)\}_{i=1}^m$ where $y_i = f(x_i)$, we apply this representation to each data point:
\begin{align*}
y_i &= f(x_i^{(1)},\ldots,x_i^{(n)})\\
&= \sum_{j=0}^{2p} \phi_j\left(\sum_{k=1}^n \lambda^k\phi(x_i^{(k)} + ja)\right)\\
&= \sum_{j=0}^{2p} g\left(z_{i,j}\right)
\end{align*}
where we've defined $z_{i,j} = \sum_{k=1}^n \lambda^k\phi(x_i^{(k)} + ja)$ and set $g = \phi_j$ for each $j$.

\cite{montanelli2020deep} proved that each outer function $\phi_j$ can be approximated by a ReLU network to arbitrary precision. Therefore, we can replace each $\phi_j$ with a single ReLU network $g$:
$$
g(x) = \sum_{k=1}^K \beta_k\text{ReLU}(w_kx + b_k)
$$
where $K$ is the number of neurons. We see that any function $f$ and its corresponding dataset can be represented as a GAM with features engineered through the K{\"o}oppen function $\phi$. \cite{lai2023kolmogorov} further developed this insight into a specialized two-layer ReLU network architecture.

\paragraph{Note} The number of outer functions grows with the input dimension $n$, not with the dataset size. The outer function $g$ is shared across all terms, making this a true GAM representation with universal feature engineering provided by the inner K\"oppen function $\phi$.

In a similar vein to \cite{hecht-nielsen1987kolmogorovs}, we show how this functional representation can be used to define a statistical model for applications in machine learning tasks \cite{breiman2001,bhadra2024}. We propose the following model:
\begin{align*}
f(x_1,\ldots,x_d) &= \sum_{q=0}^{2d} g\left(\sum_{p=1}^d \alpha_p\psi(x_p + q\alpha) + \delta_q\right)\\
g(x) &= \sum_{k=1}^K \beta_k{\rm ReLU}(w_kx + b_k)
\end{align*}
where $K$ denotes the number of neurons in the outer function's architecture. The parameters $(\beta_k, w_k, b_k)_{k=1}^K$ can be learned using $L^2$-minimization and SGD. This architecture employs only two univariate activation functions: a learned ReLU network for the outer layer and the K\"oppen function for the inner layer, which can be computed independently of $f$.

\subsection{p-adic Neural Networks and Embeddings}

A crucial aspect of KST-based architectures is their relationship to p-adic expansions and embeddings. This connection, first explored by \cite{albeverio1999memory} and recently advanced by \cite{zuniga-galindo2024hierarchical}, builds on the foundational work of \cite{chua1988cellular} in cellular neural networks of the 1980s, which encompassed cellular automata as special cases.

For any input $x$, its p-adic expansion can be written as:
\[
x = \sum_{l=1}^{\infty}i_l\gamma^{-l} = \lim_{k \rightarrow \infty} \left(\sum_{l=1}^{k}i_l\gamma^{-l}\right)
\]
where $\gamma$ serves as the base of the expansion and $i_l$ are the digits in this representation. This expansion provides a natural embedding of the input space into a higher-dimensional representation, resembling contemporary kernel embedding techniques in deep learning \cite{bhadra2024,polson2017a,breiman2001}. Just as kernel methods map inputs into high-dimensional feature spaces to capture nonlinear relationships, p-adic expansions provide a structured way to embed inputs into spaces where their topological properties become more apparent. This is also related to hashing.

\subsection{Inference}
An important property of an algorithm is how many parameters exist in the evaluation of the network, the so-called inference problem. The main advantage of our K-GAM network is that it has relatively few parameters. This also means in training that we can use efficient Bayesian learning methods.

Consider the class that corresponds to kernel  estimators of the form 
$$
f(x) = \int K( x , t ) d \mu(t) \; \; {\rm with}  \; \int d | \mu | (t) \leq K  \;  {\rm and} \; x = (x_1 , \ldots , x_d ) 
$$
We call $ K(x,t) = K_t (x) $ an $ L_\infty $-atom if $|  K_t(x) | \leq 1 $. For such estimators, there exists an M-term approximation of $f$
where C is independent of the dimension $d$ given by 
$$
\exists \; \;  f_M (x) = \sum_{j=1}^M a_j K( x , t_j  )  \; \; {\rm such \; that} 
\vnorm{f - f_N }_\infty \leq  C M^{- \frac{1}{2}  }
$$ 
Let $f_M$ denote an $M$-term approximation of $f$. \cite{bellman1961adaptive} showed that in general,  $ \vnorm{f - f_M} = O ( M^{-r/d} ) $,  where $r$ is the isotropic smoothness of $f$ and $d$ is the dimension. Hence, we encounter the curse of dimensionality. However, simple restrictions on $f$ lead to powerful results. For example, suppose  that $ \nabla f $ is in the $ L^1$-Fourier class. \cite{barron1993universal} showed that it is possible to find $f_m$ such that  $ \vnorm{f - f_M} = O ( M^{-1/2} ) $ is independent of $d$. Superpositions of Gaussians (a.k.a. radial basis functions) also have the same property.

The class of functions that we consider can be represented as $L^1 $-combinations of $ L^\infty$-atoms. For $f_M$, we will use superpositions of functions. For example, with $ t \defeq ( x_0 , s ) $ and Gaussian  atoms $ K( x , t ) = \exp \left ( - \vnorm{x - x_0 }^2 / s^2 \right ) $, see \cite{niyogi1998incorporating}. Let $ t= (x_0 , k ) $ where $k$ is an ortant indicator. Then the resultant approximation is of also of order $O ( M^{-1/2} ) $  independent of $d$. This fundamental result underlies Monte Carlo methods and CDF estimation.

A more general class is found by superposition of $2^d$ functions, each monotone for a different ortant. Then the condition $  \int d | \mu | (t) \leq 1 $ manifests as a derivative condition:
\[ 
\frac{\partial^d f }{ \partial x_1 \ldots \partial x_d } \in L_1 .
\] 
This characterizes the space of functions with bounded mixed derivatives, which naturally leads to $\beta$-H\"older smooth functions.  For such $ \beta$-H\"older continuous functions, the network typically requires $ O( M^\beta ) $ parameters (see \cite{hegland2002additive} for further results on additive models and approximation properties in high dimensions).

Kolmogorov Spline Networks, designed for differentiable functions with bounded derivatives, satisfy the stronger property:
$$
\vnorm{f - f_M} = O ( M^{- 1})
$$
with number of parameters  given by $  O( M^{2/3} ) $. This compares favorably to both the $O(M^{-1/2})$ approximation rate and the $O(M^2)$ parameter count of general one-layer hidden feed-forward networks \cite{igelnik2003kolmogorovs}.

Deep ReLU networks, using $\max(x,0)$ activation, create separating hyperplanes across $L$ layers.  Not cylinder sets (a.k.a. Trees), which are a special case.  \cite{bungartz2004sparse} shows that energy-norm based sparse grids and Kolmogorov approximation schemes have bounds independent of $d$.

\cite{khavinson1997best} shows that efficient approximation schemes can be developed by the use of superpositions.  Every continuous function of several variables can be represented by the superposition of functions with only two variables. \cite{ismailov2022three} considers discontinuous functions.

As Kolmogorov might have said: there are no true multivariate problems—only superpositions of univariate affine ones!

\subsection{Brillinger}
One of the advantages of using a single-index model is that we can use least squares to estimate the parameters. Further, we can use the Stein's lemma to derive the covariance of the estimator. \cite{brillinger2012generalized} considers the single-index model with non-Gaussian regressors where  $(Y,X) $ are stochastic with conditional distribution
$$
Y \mid X \sim  N(g (  X \beta ),~\sigma^2).
$$
Here $X \beta  $ is the single features found by data reduction from high dimensional vector $ X$.
Let vector $ \hat{\beta}_{OLS} $ denote the least squares estimator which solves $X^TY = X^TX\hat{\beta}_{OLS}$. 

By Stein's lemma (\cite{stein1981}), 
$$
\text{Cov}(g(U), V) = \text{Cov}(U, V) E(g'(U)) 
$$
if $U$ and $V$ are jointly normally distributed and $E(|g'(U)|) < \infty$. 
Hence,
$$
\frac{\text{Cov}(g(X\beta), X)}{\text{Cov}(g(X\beta), X\beta)} = \frac{\beta\text{Var}(X)}{\text{Var}(X\beta)}
$$
and
$$
\text{Cov}(Y,X)  = \beta \text{Cov} ( g (  X \beta ) ,  X \beta ) \text{Var}(X) / \text{Var}(  X \beta ) 
$$
\\
Furthermore, the Kolmogorov network enables the incorporation of linear dimensionality reduction layers, such as Partial Least Squares (PLS). Given a training dataset of input-output pairs $ ( X_i, Y_i)_{i=1}^N $, the goal is to find a prediction rule that can estimate a new output $ Y_* $ given an unseen input $ X_*$.  Let $F$ denote latent hidden features that are to be either hand-coded or learned from the data. Our nonlinear latent feature predictive model takes the form:
\begin{align*}
Y| F &  \sim p( Y| F ) \\
F & = g ( B X ) 
\end{align*}
where $ B$ is a loadings matrix that we will estimate via PLS and will provide our dimension reduction. 

Here $g$ is a data transformation that allows for relations between latent features $F = g(X)$ and $Y$ to be modeled by a well understood probabilistic model $p$. Typically, $g$ will perform dimension reduction or dimension expansion and can be learned from data. The top level of our model is stochastic. This gives a full representation of the predictive uncertainty in predicting new $ Y_* $. In many cases, our predictor is simply the conditional mean $ \hat Y_* = E(Y_*\mid F_* ) $ where $F_* = g(\hat{B}_{PLS} X_*)$.
The distinguishing feature of the model is that 
$g \defeq g_{W, b} $ is a deep learner where the parameters $ ( \hat{W} , \hat{b} ) $ are estimated using traditional stochastic gradient descent (SGD) methods.

A special case of this general framework, that provides data dimension reduction for both $Y$ and $X$ is to simply use deep learning on the score matrices $U$ and $T$. This leads to a reduced form model given by:
\[
U = G_{W,b}(T) 
\]
where $G_{W,b}$ is a differentiable deep neural network with $L$ layers. 

The optimal parameters $\hat \theta = (\hat W, \hat b)$ are estimated independently of the PLS data reduction projection using traditional SGD. Hence, the reduced form DPLS model is 
\begin{align*}
	Y &  = UQ + E\\	
	X &  = TP + F\\
	U &  = G_{W,b}(T) + \epsilon_U \label{eq:dlu}
\end{align*}
A key result due to \cite{brillinger2012generalized} and \cite{Naik2000} is that the PLS regression estimator is consistent, up to a constant factor, even when there is a non-linear relation between $U$ and $T$ as in deep learning. This process is extremely useful as we separate the two procedures. First, we apply PLS to find the data dimension reduction loadings $(P,Q)$ and the corresponding score matrices $(U,T)$. Then implement a deep neural network architecture to find the nonlinear relations $U  = G_{W,b}(T)$. 

For  classification, the DL-PLS model for the output $ Y$ which is $ n \times q $  is given by
\begin{align*}
	p( Y=1 | X ) & = f ( UQ ) \\
	U & = G_{W,b} ( T ) \\
	X & = T P + F,
\end{align*}	
where $ X  $ is $ n \times p $, the scores matrices $ T$ is $ n \times L $ and $ U $ is $ n \times L $  and $ P $ is an $ L \times p $ loadings matrix where the number of layers $ L \ll p $.

In general, we can process $X$ via the PLS transform $\hat{B}_{PLS}X$ and then directly use deep learning by modeling 
$$
Y = g(\hat{B}_{PLS}X).
$$
Uncertainty quantification for $\hat{B}_{PLS}$ is also available using asymptotically normal posterior of $B_{PLS}$ given the data.

\subsection{Kernel Smoothing: Interpolation} 
The theory of kernel methods was developed by Fredholm in the context of integral equations \cite{fredholm1903classe}. The idea is to represent a function as a linear combination of basis functions, which are called kernels. 
\[
 f(x) = \int_{a}^{b} K(x,x')  d \mu (x') dx'  \; \; {\rm where} \; \; x = ( x_1 , \ldots , x_d ) 
\]
Here, the unknown function $f(x)$ is represented as a linear combination of kernels $K(x,x')$ with unknown coefficients $\phi(x')$. The kernels are known, and the coefficients are unknown. The coefficients are found by solving the integral equation. The first work in this area was done by Abel who considered equations of the form above.

Nowadays, we call those equations Volterra integral equations of the first kind. Integral equations typically arise in inverse problems. Their significance extends beyond their historical origins, as kernel methods have become instrumental in addressing one of the fundamental challenges in modern mathematics: the curse of dimensionality.

Bartlett  \cite{nadaraya1964estimating} and \cite{watson1964smooth} proposed the use  of kernels to estimate the regression function. The idea is to estimate the regression function $f(x)$ at point $x$ by averaging the values of the response variable $y_i$ at points $x_i$ that are close to $x$. The kernel is used to define the weights.

The regression function is estimated as follows
\[
	\hat{f}(x) = \sum_{i=1}^n  y_i K(x,x_i)/ \sum_{i=1}^n K(x,x_i) ,
\]
where the kernel weights are normalized.

Both Nadaraya and Watson considered the symmetric kernel $K(x,x') = K(\|x'-x\|_2)$, where $||\cdot||_2$ is the Euclidean norm. The most popular kernel of that sort is the Gaussian kernel:
\[
	K(x,x') = \exp\left( -\dfrac{\|x-x'\|_2^2}{2\sigma^2}\right).
\]
Alternatively, the 2-norm can be replaced by the inner-product:
$K(x,x')  =  \exp\left( x^Tx'/2\sigma^2\right) $.

Later, \cite{parzen1962estimation} proposed to use kernels to estimate the density function. The idea is to estimate the density function $f(x)$ at point $x$ by averaging the values of the kernel $K(x,x_i)$ at points $x_i$ that are close to $x$. This idea was applied in many contexts by statisticians \cite{gramacy2008bayesian,higdon2008computer},  machine learners \cite{vapnik1999nature} and engineers \cite{mockus1989bayesian}.

Kernel methods are supported by numerous generalization bounds which often take the form of inequalities that describe the performance limits of kernel-based estimators. A particularly important example is the Bayes risk for $k$-nearest neighbors ($k$-NN), which can be expressed in a kernel framework as:
$$
 \hat{f} ( x) =  \sum_{i=1}^N w_i y_i        \; {\rm where} \; w_i \defeq K( x_i , x ) /  \sum_{i=1}^N K( x_i ,x )   
$$ 
$k$-NN classifiers have been proven to converge to an error rate that is bounded in relation to the Bayes error rate, with the exact relationship depending on the number of classes. For binary classification, the asymptotic error rate of $k$-NN is at most $2R^*(1-R^*)$, where $R^*$ is the Bayes error rate. This theoretical bound suggests potential for improvement in practice. Cover and Hart proved that interpolated k-NN schemes are consistent estimators, meaning that their performance improves with increasing sample size.

\paragraph{Training Rates} 
Consider the non-parametric condition regression, $ y_i= f (x_i) + \epsilon_i $ where
$ x_i = ( x_{1i} , \ldots , x_{di} ) $. We wish to estimate $ f( x_1 , \ldots , x_d ) $ 
where $ x  = ( x_1 , \ldots , x_d ) \in [0,1]^d $. From a classical risk perspective, define
$$
R ( f , \hat{f}_N ) = E_{X,Y} \left ( \lVert  f - \hat{f}_N \rVert^2 \right ) 
$$
where $ \lVert . \rVert $ denotes $ L^2 ( P_X) $-norm.

Under standard assumptions, we have an optimal minimax rate $ \inf_{\hat{f}} \sup_f R( f , \hat{f}_N ) $ of
$ O_p \left ( N^{- 2 \beta /( 2 \beta + d )} \right ) $ for $\beta$-H\"older smooth functions $f$.
This rate still depends on the dimension $d$, which can be problematic in high-dimensional settings. By restricting the class of functions, better rates can be obtained, including ones that do not depend on $d$. In this sense, we avoid the curse of dimensionality. Common approaches include considering the class of linear superpositions (a.k.a. ridge functions) and projection pursuit models.

Another asymptotic result comes from a posterior concentration property. Here, $ \hat{f}_N $ is constructed as a regularized  MAP (maximum a posteriori) estimator, which solves the optimization problem
$$
\hat{f}_N = \arg \min_{ \hat{f}_N } \frac{1}{N} \sum_{i=1}^N ( y_i - \hat{f}_N ( x_i )^2 + \phi ( \hat{f}_N ) 
$$
where $\phi(\hat{f})$ is a regularization term. Under appropriate conditions, the ensuing posterior distribution $ \Pi(f | x, y) $ can be shown to concentrate around the true function at the minimax rate (up to a $\log N$ factor).

A key result in the deep learning literature provides convergence rates for deep neural networks. Given a training dataset of input-output pairs $ ( x_i , y_i)_{i=1}^N $ from the model
$ y = f(x) + \epsilon $ where $f$ is a deep learner (i.e. superposition of functions
$$
f = g_L \circ \ldots g_1 \circ g_0
$$
where each $g_i$ is a $ \beta_i$-smooth H\"older function with $ d_i $ variables, that is $ | g_i (x) -g_i (y) < | x-y |^\beta $.

Then, the estimator has optimal rate:
$$
O \left ( \max_{1\leq i \leq L } N^{- 2 \beta^* /( 2 \beta^* + d_i ) } \right )  \; {\rm where} \; \beta_i^* = \beta_i \prod_{l = i+1}^L \min ( \beta_l , 1 ) 
$$
This result can be applied to various function classes, including generalized additive models of the form
$$
f_0 ( x ) = h \left ( \sum_{p=1}^d f_{0,p} (x_p) \right )
$$
where 
$ g_0(z) = h(z) $, $g_1 ( x_1 , \ldots , x_d ) = ( f_{01}(x_1) , \ldots , f_{0d} (x_d) ) $  and $ g_2 ( y_1 , \ldots , y_d ) = \sum_{i=1}^d y_i  $. In this case, $d_1 = d_2 = 1$, and assuming $h$ is Lipschitz, we get an optimal rate of $O(N^{-1/3})$, which is independent of $d$.

\cite{schmidt-hieber2021kolmogorov} show that deep ReLU networks also have optimal rate of $ O( N^{-1/3} ) $ for certain function classes. For $3$-times differentiable (e.g. cubic B-splines ), \cite{coppejans2004kolmogorovs} finds a rate of $ O( N^{-3/7} ) = O( N^{-3/(2 \times 3 + 1) } ) $. \cite{igelnik2003kolmogorovs} finds a rate $ O( N^{-1} ) $ for Kolmogorov Spline Networks.

Finally, it's worth noting the relationship between expected risk and empirical risk. The expected risk, $R$, is typically bounded by the empirical risk plus a term of order $1/\sqrt{N}$:
$$
R(y, f^\star) \leq \frac{1}{N} \sum_{i=1}^N R(y_i, f^\star(x_i)) + O\left(\frac{\|f\|}{\sqrt{N}}\right)
$$
where $f^\star$ is the minimizer of the expected risk. However, in the case of interpolation, where the model perfectly fits the training data, the empirical risk term becomes zero, leaving only the $O(1/\sqrt{N})$ term.

\section{Transformers as Kernel Smoothing}
\cite{bahdanau2014neural} proposed using kernel smoothing for sequence-to-sequence learning. This approach estimates the probability of the next word in the sequence using a so-called context vector, which is a weighted average of the vectors from the input sequence $h_j$:
\[
	c_i = \sum_{j=1}^n \alpha_{ij} h_j,
\]
where $\alpha_{ij}$ are the weights. The weights are defined by the kernel function:
\[
	\alpha_{ij} = \dfrac{\exp\left( e_{ij}\right)}{\sum_{k=1}^n \exp\left( e_{ik}\right)}.
\]
Instead of using a traditional similarity measure like the 2-norm or inner product, the authors used a neural network to define the energy function $e_{ij} = a(s_{i-1},h_j)$. This neural network measures the similarity between the last generated element of the output sequence $s_{i-1}$ and $j$-th element of the input sequence $h_j$. The resulting context vector is then used to predict the next word in the sequence.

\subsection{Transformer}

Transformers have since become a main building block for various natural language processing (NLP) tasks and has been extended to other domains as well due to their effectiveness. The transformer architecture is primarily designed to handle sequential data, making it well-suited for tasks such as machine translation, language modeling, text generation, and more. It achieves state-of-the-art performance by leveraging a novel attention mechanism.

The idea to use kernel smoothing for sequence to sequence was called ``attention'', or cross-attention, by \cite{bahdanau2014neural}. When used for self-supervised learning, it is called self-attention. When a sequence is mapped to a matrix $M$, it is called multi-head attention. The concept of self-attention and attention for natural language processing was further developed by \cite{vaswani2023attention} who developed a smoothing method that they called the transformer.

The transformer architecture revolves around a series of mathematical concepts and operations:

\begin{itemize}
\item Embeddings: The input text is converted into vectors using embeddings. Each word (or token) is represented by a unique vector in a high-dimensional space.
\item Positional Encoding: Since transformers do not have a sense of sequence order (like RNNs do), positional encodings are added to the embeddings to provide information about the position of each word in the sequence.
\item Multi-Head Attention: The core of the transformer model. It enables the model to focus on different parts of the input sequence simultaneously. The attention mechanism is defined as:
  $$ \text{Attention}(Q, K, V) = \text{softmax}\left(\frac{QK^T}{\sqrt{d_k}}\right)V $$
  where $Q$, $K$, and $V$ are query, key, and value matrices respectively.
\item Query (Q), Key (K), and Value (V) Vectors: These are derived from the input embeddings. They represent different aspects of the input. 
\item Scaled Dot-Product Attention: The attention mechanism calculates the dot product of the Query with all Keys, scales these values, and then applies a softmax function to determine the weights of the Values.
\item Multiple 'Heads': The model does this in parallel multiple times (multi-head), allowing it to capture different features from different representation subspaces. 
\item Layer Normalization and Residual Connections: After each sub-layer in the encoder and decoder (like multi-head attention or the feed-forward layers), the transformer applies layer normalization and adds the output of the sub-layer to its input (residual connection). This helps in stabilizing the training of deep networks.
\item Feed-Forward Neural Networks: Each layer in the transformer contains a fully connected feed-forward network applied to each position separately and identically. It is defined as:
  $$ \text{FFN}(x) = \max(0, xW_1 + b_1)W_2 + b_2 $$
  where $W_1$, $W_2$, $b_1$, and $b_2$ are learnable parameters.
\item Output Linear Layer and Softmax: The decoder's final output passes through a linear layer followed by a softmax layer. This layer converts the decoder output into predicted next-token probabilities.
\item Training and Loss Function: Transformers are often trained using a variant of Cross-Entropy Loss to compare the predicted output with the actual output.
\item Masking: In the decoder, to prevent future tokens from being used in the prediction, a technique called 'masking' is applied.
\item Backpropagation and Optimization: The model's parameters are adjusted through backpropagation and optimization algorithms like Adam.
\end{itemize}

Later, \cite{lin2017structured} proposed using similar idea for self-supervised learning, where a sequence of words (sentence) is mapped to a single matrix:
\[
	M = AH,
\] 
where $H$ is the matrix representing an input sequence $H = (h_1,\ldots,h_n)$ and $A$ is the matrix of weights:
\[
	A = \mathrm{softmax}\left(W_2\tanh\left(W_1H^T\right)\right).
\]
This allows to represent a sequence of words of any length $n$ using a ``fixed size'' $r\times u$ matrix $M$, where $u$ is the dimension of a vector that represents an element of a sequence (word embedding) and $r$ is the hyper-parameter that defines the size of the matrix $M$.

The main advantage of using smoothing techniques (transformers) is that they are parallelizable. Current language models such as BERT, GPT, and T5 rely on this approach. Further, they also has been applied to computer vision and other domains. Its ability to capture long-range dependencies and its scalability have made it a powerful tool for a wide range of applications. See \cite{tsai2019transformer} et al for further details.

\section{Application}

\subsection{Simulated Data}
\begin{figure}
	\centering
	\includegraphics[width=0.5\textwidth]{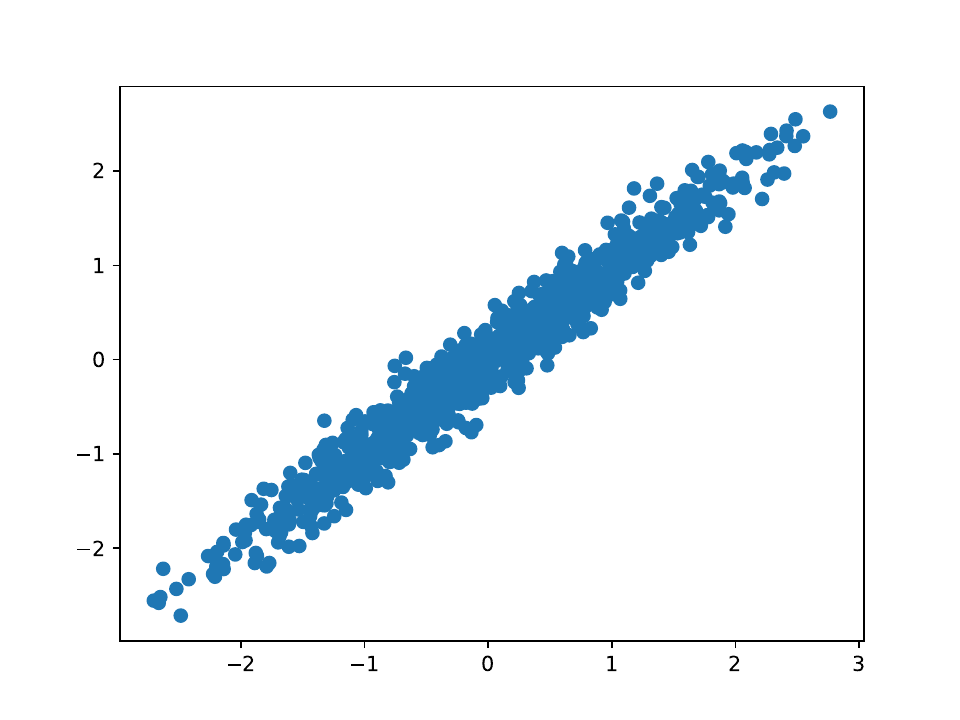}
	\caption{Scatter plot of the simulated dataset}\label{fig:simdata-scatter}
\end{figure}

We also apply the K-GAM architecture to a simulated dataset to evaluate its performance on data with known structure and relationships. The dataset contains 100 observations generated from the following function:
\begin{align*}
	&y = \mu(x) + \epsilon, \quad \epsilon \sim \mathcal{N}(0,1)\\
	&\mu(x) = 10\sin(\pi x_1 x_2) + 20(x_3-0.5)^2 + 10x_4 + 5x_5.
\end{align*}
The goal is to predict the function $y(x)$ based on the input $x$. The dataset is often used as a benchmark dataset for regression algorithms due to its diverse mix of relationships (linear, quadratic, nonlinear, Gaussian random noise) between the input features and the target function. The plot of $\mu$ (no noise) vs $y$ (noise) is shown in Figure \ref{fig:simdata-scatter}.

We use the K\"oppen function to transform the five-dimensional input into a set of 11 features ($2d+1$). We then learn the outer function $g$ using a ReLU network. To thoroughly investigate the model's capabilities, we implement two distinct approaches to learning the outer function. The first approach uses different $g$ functions for each feature, following the original KST formulation. This allows each function to specialize in capturing specific patterns, but might be more difficult to train and has more parameters. The second approach uses a single $g$ function for all features, as proposed by \cite{lorentz197613th}, providing a more unified and parameter-efficient representation.

\begin{figure}
\begin{tabular}{ccc}
	\includegraphics[width=0.33\textwidth]{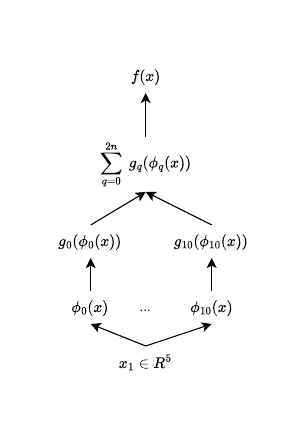} & \includegraphics[width=0.30\textwidth]{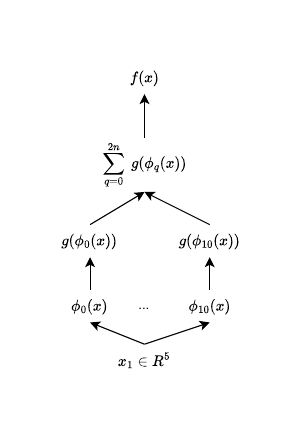}& \includegraphics[width=0.30\textwidth]{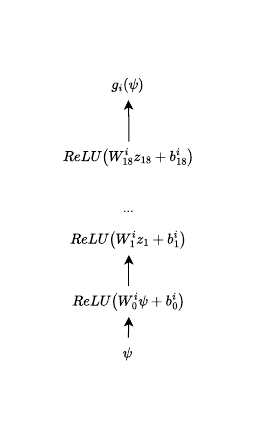}\\
	(a) Multiple $g_i$ functions& (b) Single $g$ function & (c) $g$ function architecture
\end{tabular}
\caption{KST architecture for the simulated dataset}\label{fig:outer-gfunc}
\end{figure}

\begin{figure}[h]
\begin{tabular}{cc}
	\includegraphics[width=0.5\textwidth]{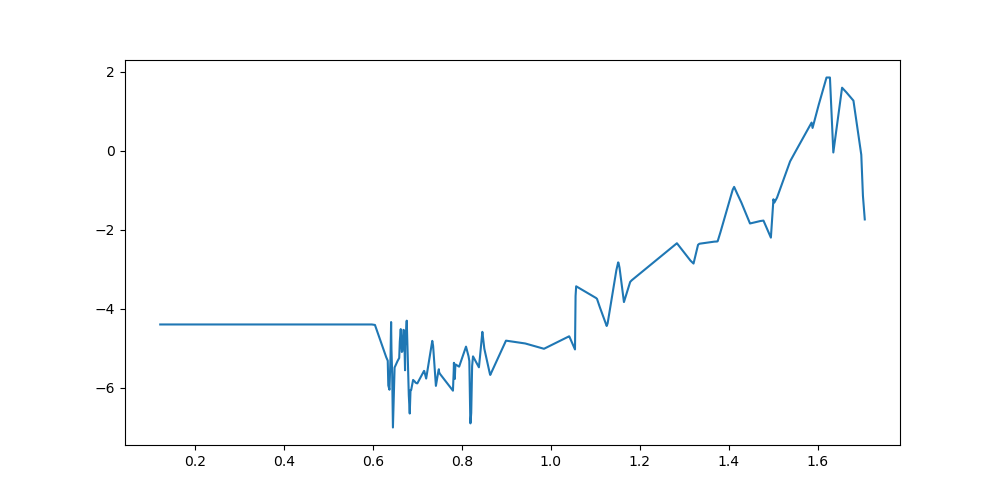} & \includegraphics[width=0.5\textwidth]{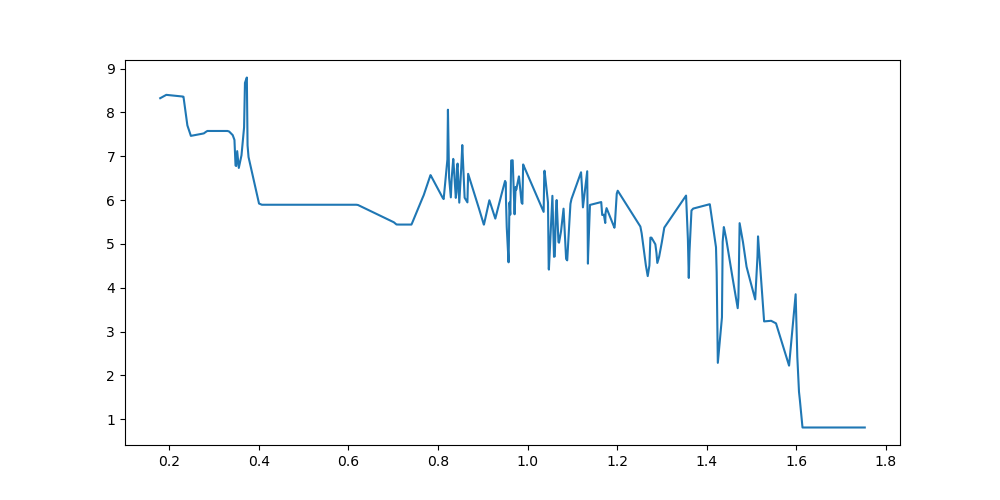}\\ 
	(a) $g_0$ & (b) $g_2$\\
	\includegraphics[width=0.5\textwidth]{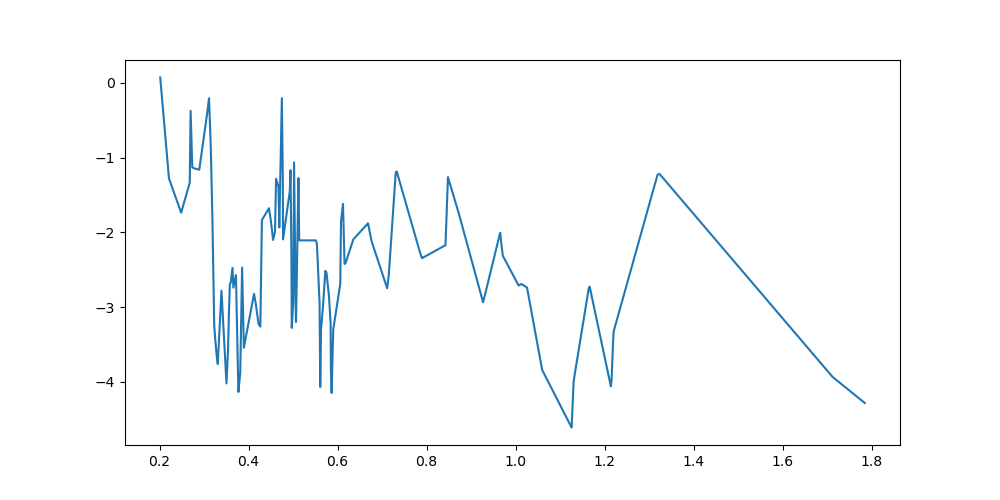} & \includegraphics[width=0.5\textwidth]{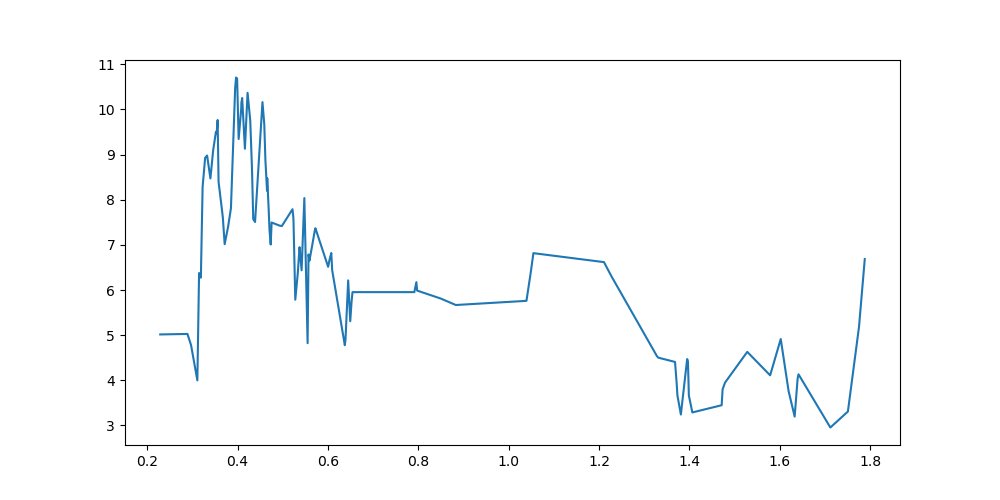}\\
	(c) $g_6$ & (d) $g_8$\\
\end{tabular}
\caption{Examples of outer functions $g_0,g_2,g_6,g_8$  for the simulated dataset}\label{fig:g2-feature}
\end{figure}

Figure \ref{fig:outer-gfunc} illustrates these architectural choices in detail, showing how the information flows through each version of the model. For the first model with multiple $g_i$ functions, the dimensions of each $g_i$ are as follows: $W^0_i \in \mathbb{R}^{16\times 1}$ and for $j=1,\ldots,18$, $W^j_i \in \mathbb{R}^{16\times 16}$. Exemplary outer functions are shown in Figure \ref{fig:g2-feature}.

\begin{figure}[H]
	\centering
	\includegraphics[width=0.6\textwidth]{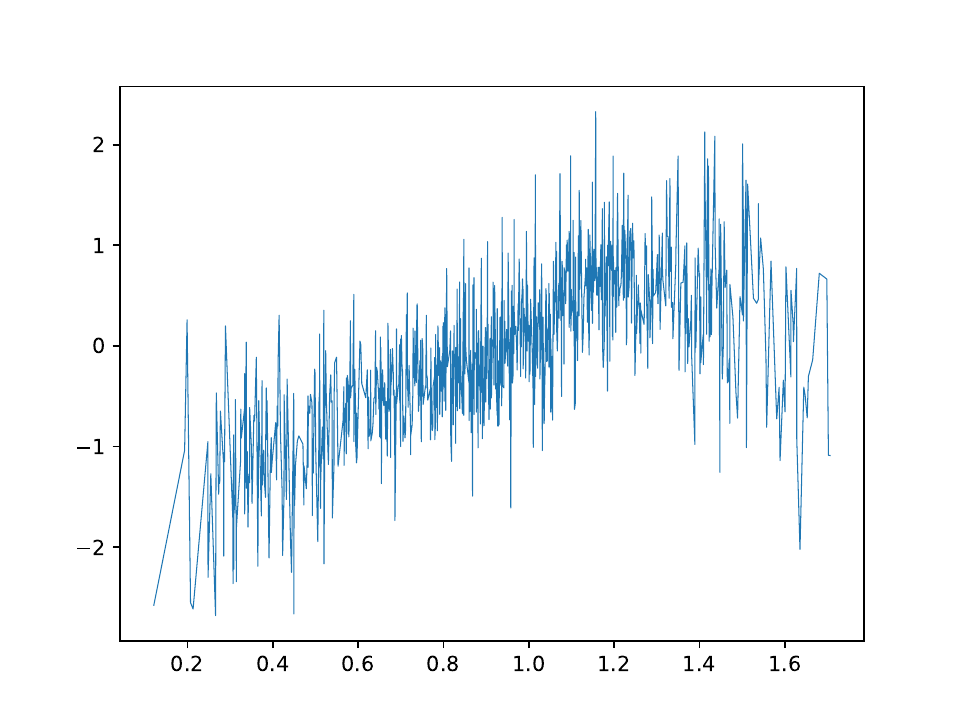}
	\caption{Plot of the single outer function $g$ for the simulated dataset}\label{fig:g1}
\end{figure}

The next architecture, which used only one function $g$ for all features, maintains a similar structure to the multiple $g$ functions approach. The only difference is in the dimensionality of the inner layers: we increased the width from 16 to 200. This increased capacity allows the single function to learn more complex patterns and compensate for the constraint of using just one function instead of multiple specialized ones. The behavior of this unified outer function is shown in Figure \ref{fig:g1}, where we can observe how it adapts to handle multiple feature transformations simultaneously.

\subsection{Iris Daa}
\begin{figure}[H]
\begin{tabular}{cc}
	\includegraphics[width=0.5\textwidth]{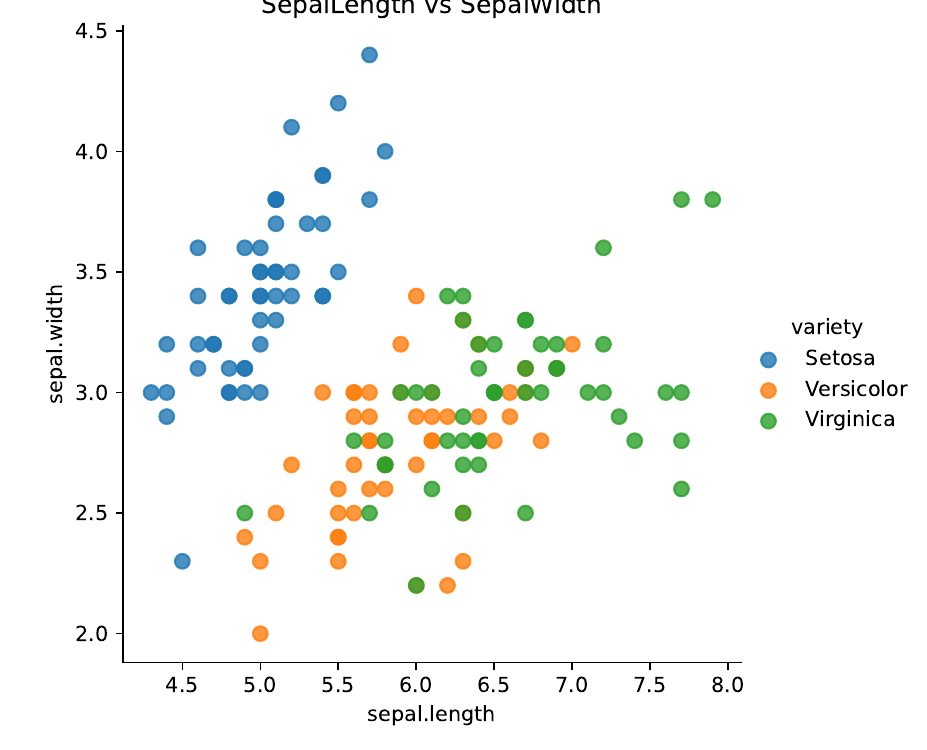} & \includegraphics[width=0.5\textwidth]{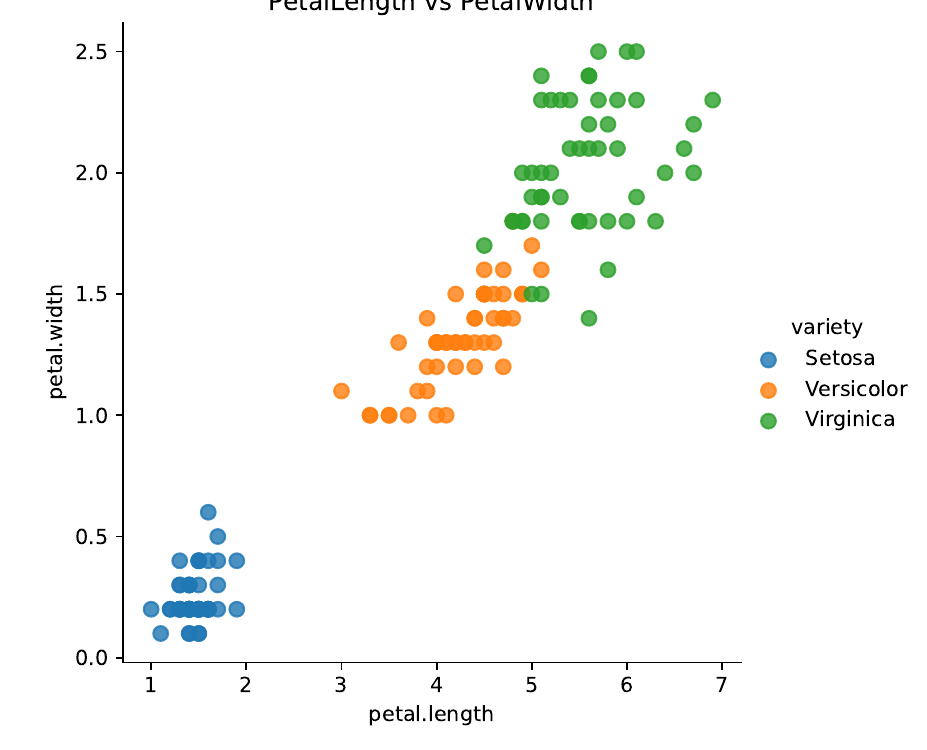}\\
	(a) Sepal Length vs Sepal Width & (b) Petal Length vs Petal Width\\
	\includegraphics[width=0.5\textwidth]{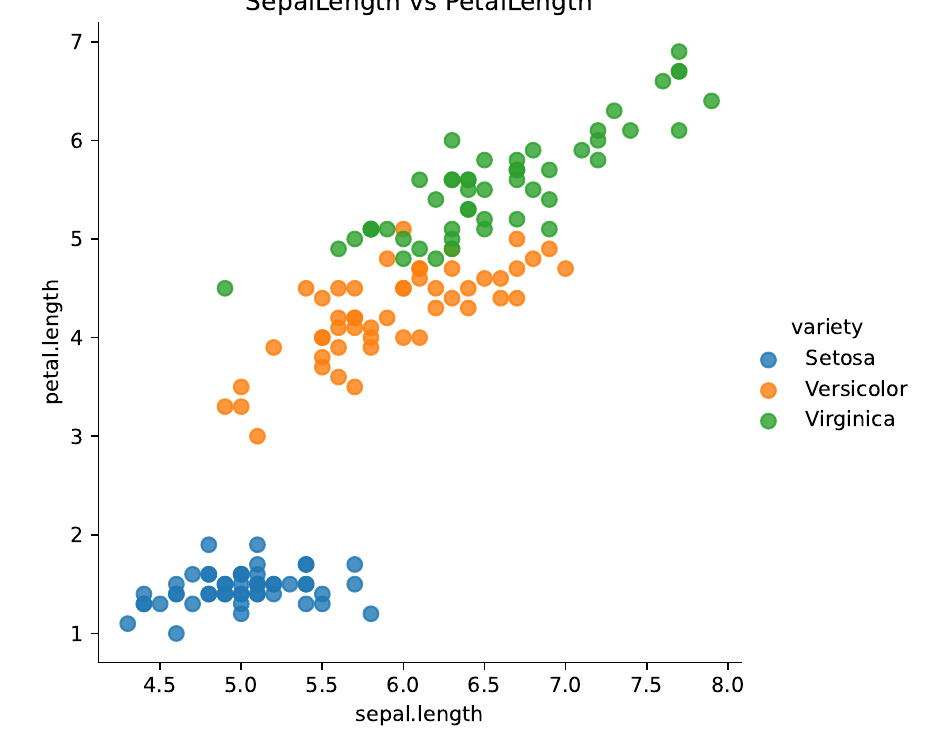} & \includegraphics[width=0.5\textwidth]{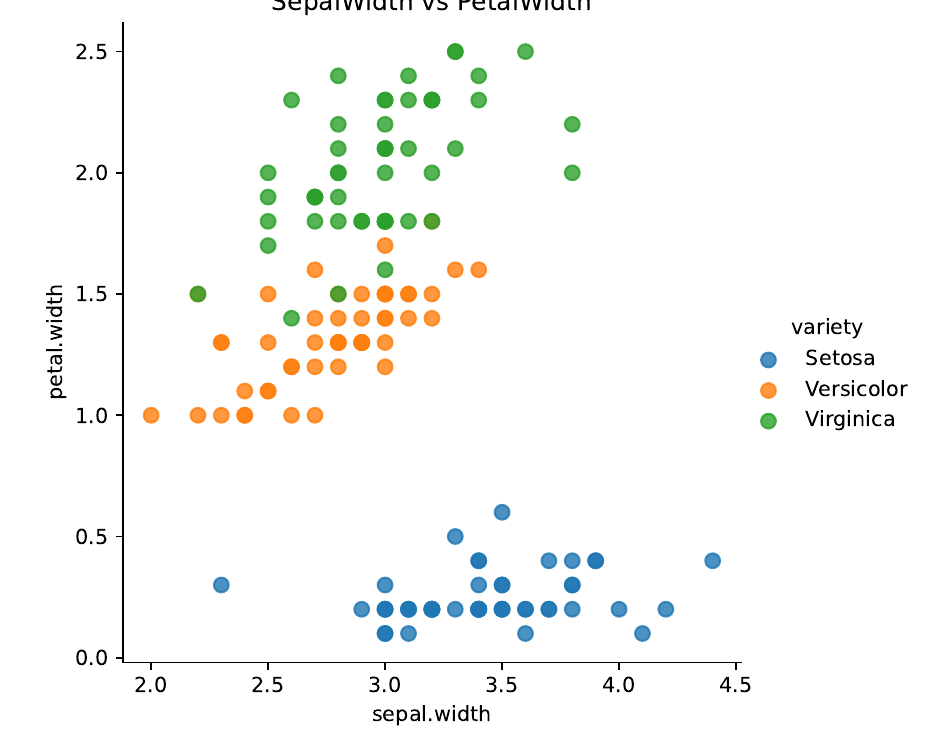}\\
	(c) Sepal Length vs Petal Length & (d) Sepal Width vs Petal Width
\end{tabular}
\caption{Scatter plots of the iris dataset}\label{fig:iris-scatter}
\end{figure}

We apply the KST architecture to an iris dataset. The iris dataset is a classic dataset in machine learning and statistics. It contains 150 observations of iris flowers. Each observation contains four features: sepal length, sepal width, petal length, and petal width. The goal is to predict the species of the iris flower based on these features. The dataset contains three classes of iris flowers: setosa, versicolor, and virginica. The dataset is often used as a benchmark dataset for classification algorithms. The dataset has 5 variables, which include four characteristics of the iris flower and the species of the flower. Figure \ref{fig:iris-scatter} shows the scatter plots of the iris dataset.

We calculate the mean $\mu_{\mathrm{SL}}$ of the  Sepal Length column  and use binary variable $y_i = \text{Sepal Length}_i > \mu_{\mathrm{SL}}$ as the output. We use the other three flower characteristics $x_1,x_2,x_3$ as input variables. We used classical GAM model to fit the data and compared it to the KST-GAN. The classibal generalized additive model  is given by
\[
\text{logit}(P(y=1)) = \beta_0 + f_1(x_2) + f_2(x_3) + f_3(x_4),
\]
where $f_1,f_2,f_3$ are smooth functions of the input features. The KST-GAN model is given by
\[
\text{logit}(P(y=1)) = \beta_0 + f_1(\psi_1)+\ldots+f_7(\psi_7),
\]
where $\psi_q$ is the K\"oppen tanformation function of the input features
\[
\phi_q(x_1,x_2,x_3) = \sum_{i=1}^3\alpha_i \psi(x_i+aq), q = 1,\ldots,7.
\]
We use $m = 2d+1 = 7$ and $k=6$ to transform the input features into a set of 7 features. We then learn the outer function $g$ using a classical GAN approach. We used the \verb|mgcv| package in R to fit the GAM model. This packages uses a penalized likelihood approach to fit the model \cite{wood2017generalized}.

Table \ref{tab:iris} comprares the performance of the GAM and KST-GAM models. We also include a classical logistic regression model for comparison. The KST-GAM model has a higher AIC and BIC compared to the GAM model. The KST-GAM model has a comparable RMSE.
\begin{table}[H]
\centering
\begin{tblr}[         
]                     
{                     
colspec={Q[]Q[]Q[]Q[]},
column{2,3,4}={}{halign=c,},
column{1}={}{halign=l,},
hline{10}={1,2,3,4}{solid, black, 0.05em},
}                     
\toprule
& Classical GAM & KST GAM & GLM \\ \midrule 
(Intercept)  & \num{-1.758}  & \num{18.064}   & \num{-28.339} \\
& (\num{1.435}) & (\num{54.091}) & (\num{8.371}) \\
Sepal.Width  &                &                 & \num{2.669}   \\
&                &                 & (\num{1.499}) \\
Petal.Length &                &                 & \num{6.377}   \\
&                &                 & (\num{2.090}) \\
Petal.Width  &                &                 & \num{-5.105}  \\
&                &                 & (\num{2.292}) \\
Num.Obs.     & \num{105}     & \num{105}      & \num{105}     \\
R2           & \num{0.790}   & \num{0.591}    &                \\
AIC          & \num{43.0}    & \num{207.3}    & \num{50.6}    \\
BIC          & \num{59.3}    & \num{258.1}    & \num{61.2}    \\
Log.Lik.     &                &                 & \num{-21.281} \\
RMSE         & \num{0.22}    & \num{0.29}     & \num{0.26}    \\
\bottomrule
\end{tblr}
\caption{Summary of the classical GAM, KST GAM, and GLM models applied to the iris dataset}\label{tab:iris}
\end{table} 

Tables \ref{tab:confusiongam} and \ref{tab:confusion} show the confusion matrices for the GAM and KST GAM models, respectively. The KST GAM model has a lower accuracy compared to the GAM model. 

\begin{table}[H]
\centering
\begin{tabular}{c|cc}
& Predicted 0 & Predicted 1\\\hline
Actual 0 & 21 & 1\\
Actual 1 & 2 & 21
\end{tabular}
\caption{Confusion out-of-sample matrix for GAM model applied to the iris dataset }\label{tab:confusiongam}
\end{table}

\begin{table}[H]
\centering
\begin{tabular}{c|cc}
& Predicted 0 & Predicted 1\\\hline
Actual 0 & 19 & 3\\
Actual 1 & 8 & 15
\end{tabular}
\caption{Confusion out-of-sample matrix for KST GAM model applied to the iris dataset }\label{tab:confusion}
\end{table}

\begin{figure}[H]
\begin{tabular}{cc}
	\includegraphics[width=0.5\textwidth]{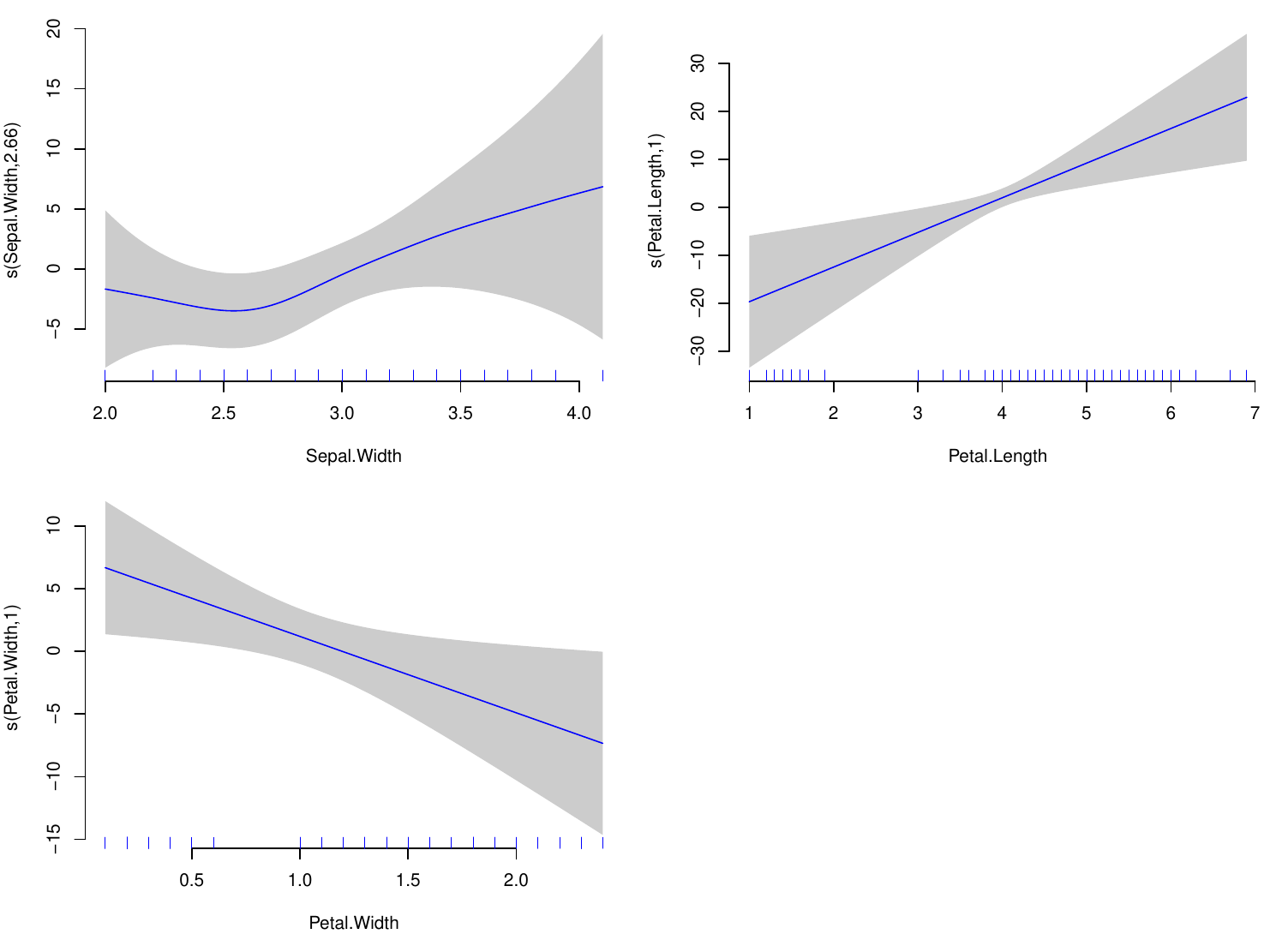} & \includegraphics[width=0.5\textwidth]{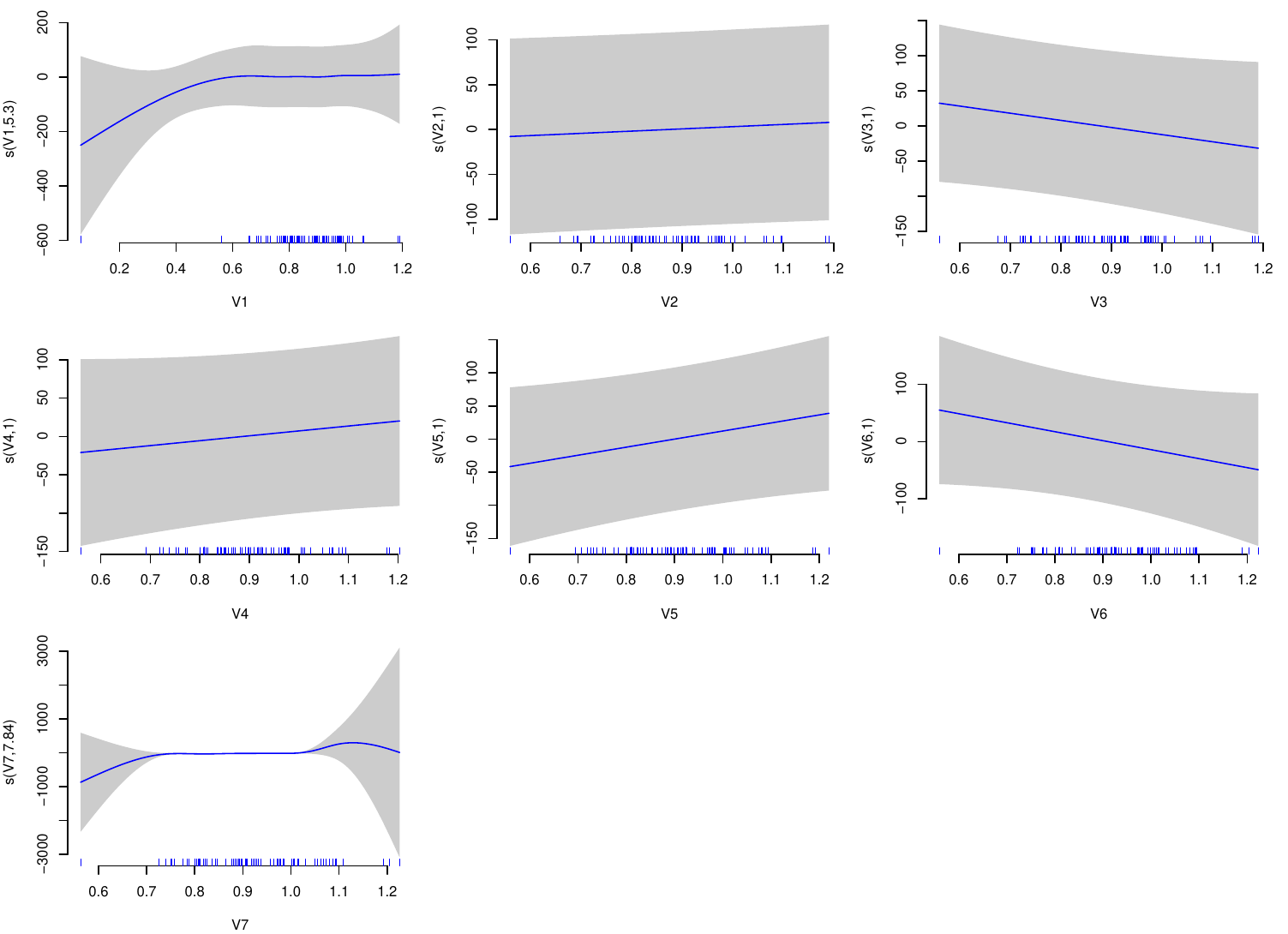}\\
	(a) GAM & (b) KST
\end{tabular}
\caption{Generalized additive model for the iris dataset versus KST.}\label{fig:gam}
\end{figure}

Figure \ref{fig:gam} shows the original features ploted against those fitted by the GAM function for both original inputs and the KST transformed inputs. 

Overall, the iris example demonstrate the inability of the GAM model to capture the complex relationships between the tranformed features and the target variable. 

\section{Discussion}
The application of K-GAM to the iris dataset reveals interesting patterns. Figure \ref{fig:psi} shows the K\"oppen function's behavior on the iris dataset. It displays a stair-step pattern that illuminates a key feature of the K\"oppen function, which is its ability to partition the input space.

The comparison between traditional GAM and K-GAM approaches in Figure \ref{fig:gam} shows some notable differences. The GAM plots show slightly smoother relationships between individual features and the output. For instance, petal width and petal length show particularly strong linear relationships with the target variable. In comparison, the K-GAM architecture seems to capture different aspects of the feature relationships. Notably, while one of the features retains a largely linear relationship, the other two features encode more complicated relationships. This difference demonstrates the effect of the K\"oppen function's mapping. 

Next, the simulated data study explores both how K-GAM handles data with known structure and the differences in using multiple outer functions versus one shared outer function within the K-GAM. The comparison between multiple $g_i$ functions (Figure \ref{fig:g2-feature}) and a single $g$ function (Figure \ref{fig:g1}) demonstrates the architectural flexibility of the approach. The single $g$ function variant shows a highly variable pattern across its input range, showing that the increased dimension compensates for the reduced flexibility of having a single function by developing more complex internal representations. The ability to choose between multiple specialized functions or a single shared function provides a useful degree of freedom in model design.

These results demonstrate that while K-GAM can effectively model both real (iris) and simulated data, the internal representations it learns may be more complex than traditional GAM approaches. The approach appears particularly effective at capturing nonlinear patterns, though the interpretability of individual feature effects becomes more challenging due to the K\"oppen function transformation. Importantly, the K-GAM approach requires significantly fewer parameters compared to standard GAMs, as it leverages a shared embedding space through the K\"oppen function as it can capture nonlinear relationships without requiring explicit interaction terms.

Our findings suggest several promising directions for future research across both theoretical and practical domains. A priority is the enhancement of the scalability of KST-based approaches along with the characterization of the function classes for which K-GAM performs optimally over existing alternatives. An important optimizaiton for K-GAM would be a specialized optimization algorithm capable of handling the discontinuities inherent in the K\"oppen function and the development of an efficient training algorithm specifically designed for high-dimensional problems.

These findings suggest that K-GAM networks represent a promising direction for efficient function approximation, particularly in scenarios where traditional deep learning approaches may be computationally intractable or parameter-inefficient. The combination of theoretical guarantees from Kolmogorov's superposition theorem with modern machine learning techniques opens new avenues for both theoretical research and practical applications.

\bibliography{KolmogorovSprecher,ref} 
\end{document}